\documentclass[letterpaper, 10 pt, conference]{ieeeconf}  

\IEEEoverridecommandlockouts                              

\usepackage{graphics} 
\usepackage{epsfig} 
\usepackage{mathptmx} 
\usepackage{times} 
\usepackage{amsmath} 
\usepackage{amssymb}  
\usepackage{hyperref}

\title{\LARGE \bf
HMCF: A Human-in-the-loop Multi-Robot Collaboration Framework Based on Large Language Models
}

\author{Zhaoxing Li, Wenbo Wu, Yue Wang, Yanran Xu, William Hunt, and Sebastian Stein
 \thanks{This work was supported by the UK Engineering and Physical
Sciences Research Council (EPSRC) through a Turing AI Fellowship
(EP/V022067/1) on Citizen-Centric AI Systems (https://ccais.ac.uk/). \newline All the authors are with the School of Electronics and Computer Science, University of Southampton, Southampton, U.K.}
}

\begin{document}

\maketitle
\thispagestyle{empty}
\pagestyle{empty}

\begin{abstract}
Rapid advancements in artificial intelligence (AI) have enabled robots to perform complex tasks autonomously with increasing precision. However, multi-robot systems (MRSs) face challenges in generalization, heterogeneity, and safety, especially when scaling to large-scale deployments like disaster response. Traditional approaches often lack generalization, requiring extensive engineering for new tasks and scenarios, and struggle with managing diverse robots. To overcome these limitations, we propose a Human-in-the-loop Multi-Robot Collaboration Framework (HMCF) powered by large language models (LLMs). LLMs enhance adaptability by reasoning over diverse tasks and robot capabilities, while human oversight ensures safety and reliability, intervening only when necessary. Our framework seamlessly integrates human oversight, LLM agents, and heterogeneous robots to optimize task allocation and execution. Each robot is equipped with an LLM agent capable of understanding its capabilities, converting tasks into executable instructions, and reducing hallucinations through task verification and human supervision. Simulation results show that our framework outperforms state-of-the-art task planning methods, achieving higher task success rates with an improvement of 4.76\%. Real-world tests demonstrate its robust zero-shot generalization feature and ability to handle diverse tasks and environments with minimal human intervention. 
\end{abstract}

\section{Introduction}

The rapid progress of artificial intelligence (AI) and robotics has significantly advanced multi-robot systems (MRSs), enabling them to perform increasingly complex tasks with high autonomy and precision~\cite{mitchell2022review}. These systems are deployed in various real-world applications, including disaster response, industrial automation, logistics, and healthcare~\cite{ghassemi2022multi, wu2024proof}. By coordinating multiple robots, MRSs can efficiently handle large-scale tasks that exceed the capabilities of individual robots.

Despite these advancements, several key challenges hinder the effective deployment of MRSs in real-world environments. First, generalization remains a major obstacle—existing methods often rely on task-specific algorithms that require extensive re-engineering when applied to new tasks or environments~\cite{chen2023agentverse}. Second, heterogeneity among robots complicates coordination—robots with varying hardware capabilities, sensor configurations, and control mechanisms must seamlessly collaborate while maintaining system-wide efficiency~\cite{chen2024scalable}. Third, existing large language models (LLMs) have shown promise in robot control, but their direct application to MRSs remains underexplored~\cite{mandi2024roco}. While LLMs can interpret high-level commands and generate task plans, they are prone to hallucinations, producing infeasible or unsafe instructions that can compromise real-world deployments~\cite{ye2023cognitive}.

To address these challenges, we introduce the \underline{H}uman-in-the-loop \underline{M}ulti-robot \underline{C}ollaboration \underline{F}ramework (HMCF), an innovative LLM-powered framework designed for dynamic and efficient multi-robot collaboration. Our framework combines LLM-based task allocation, real-time human oversight, and decentralized task verification to enable scalable and adaptive multi-robot coordination. In HMCF, each robot operates with an LLM-powered agent that understands its own capabilities, translates tasks into executable instructions, and verifies commands before execution. A central assistant LLM agent manages task assignments, while a human-in-the-loop mechanism provides oversight and intervention when needed. This hybrid approach ensures higher adaptability, safety, and robustness in diverse environments.

The key contributions of this paper are as follows:
\begin{itemize}
    \item \textbf{A novel LLM-powered multi-robot collaboration framework (HMCF)}: We develop a human-in-the-loop multi-robot coordination framework that enables efficient and scalable task allocation among heterogeneous robots using LLM-based reasoning.
    \item \textbf{LLM-assisted generalization across diverse tasks and environments}: Unlike traditional task-specific multi-agent collaboration methods, HMCF leverages LLMs to generalize across diverse tasks, reducing the need for extensive task-specific customization.
    \item \textbf{A user-friendly human-robot interaction interface}: We design an intuitive interface that allows users to configure, monitor, and intervene in multi-robot operations using natural language commands, reducing control complexity and operational costs.
    \item \textbf{LLM hallucination mitigation via human-in-the-loop and task verification}: Our framework integrates LLM agents that verify task feasibility before execution and enables human intervention when necessary, ensuring safe and reliable multi-robot collaboration.
\end{itemize}

To evaluate the effectiveness of our framework, we conduct a two-part experimental study. First, we perform a simulated evaluation using the BEHAVIOR-1K benchmark, comparing HMCF against five state-of-the-art multi-robot collaboration methods in diverse environments, which demonstrates a 4.76\% improvement in task success rates. Second, we conduct a real-world deployment with heterogeneous robots, validating HMCF’s adaptability and robustness using a team of wheeled and legged robots, showcasing its zero-shot generalization capabilities and seamless human-robot collaboration.

The remainder of this paper is organized as follows: Section~\ref{LR} reviews related work on LLMs for MRSs and robotics. Section~\ref{method} details the HMCF framework. Section~\ref{exp} presents experiments and evaluations in simulated and real-world environments. Section~\ref{dis} discusses experimental findings, limitations, and future directions, and Section~\ref{con} concludes the paper.

\section{Related Work} \label{LR}
The integration of Large Language Models (LLMs) into multi-robot systems (MRSs) has shown promise in enhancing collaboration, generalization, and scalability in dynamic environments~\cite{kim2024survey,li2021survey,li2024towards}. LLMs, originally developed for language processing, now facilitate decision-making, task execution, and communication in multi-agent systems (MASs)~\cite{wang2024survey, guo2024large}. Their ability to process natural language enables seamless interaction between human users and robots~\cite{liang2020emergent,li2023behavior}. However, key challenges persist in task allocation, execution safety, and hallucination mitigation~\cite{mozannar2024effective}.

In robot control, LLMs leverage zero-shot and few-shot learning for adaptability in unstructured environments~\cite{kim2024survey,li2023broader}. By integrating vision transformers, robots can map visual inputs to actions, improving perception and multi-modal task execution~\cite{brohan2023rt,li2023deep}. LLMs also enhance high-level task planning, using frameworks like planning domain definition languages for structured decision-making~\cite{shen2024language,li2023sim,li2023towards}. However, their lack of a deep world model leads to errors in multi-step planning~\cite{valmeekam2024planbench,li2024design}.
For multi-robot collaboration, LLMs help address communication and task allocation challenges. Systems like SMART-LLM enable natural language-based coordination, improving flexibility in heterogeneous teams~\cite{kannan2023smart,li2023broader,li2024integrating,li2024lbkt,li2023behavior,li2021survey}. DMRS-2D further decentralizes decision-making through iterative communication~\cite{zhang2023building}, while Roco integrates LLM-driven collision-free cooperation and human-robot teaming~\cite{mandi2024roco}. However, most approaches only incorporate humans at the task-setting stage, limiting their involvement during execution, where LLM hallucinations can introduce critical errors and risks~\cite{martino2023knowledge}.

Human-in-the-loop (HITL) systems offer a solution by enabling real-time human oversight and intervention, and ensuring compliance with safety and performance criteria~\cite{hussein2018mixed}. HITL integration into LLM-driven MRSs allows natural language command and control, reducing cognitive load and improving task alignment with human expectations~\cite{li2021survey}. These mechanisms also support humans as active team members, participating in planning and execution alongside robots~\cite{zhang2023building}.


\section{Methodology} \label{method}

In this section, we first present the overall framework design of HMCF, outlining how tasks are assigned, verified, and dynamically reallocated using two kinds of LLM agents to ensure seamless collaboration. Next, we describe the human-robot interactive interface, which enables intuitive robot management and task verification. We then explain the roles of LLM agents and human users in facilitating robot interactions, reducing hallucinations, and improving task execution efficiency. Finally, we discuss robot deployment cases in both simulation and real-world scenarios.

\subsection{Framework Design}

\begin{figure}[t]
  \centering
  \includegraphics[width=\linewidth]{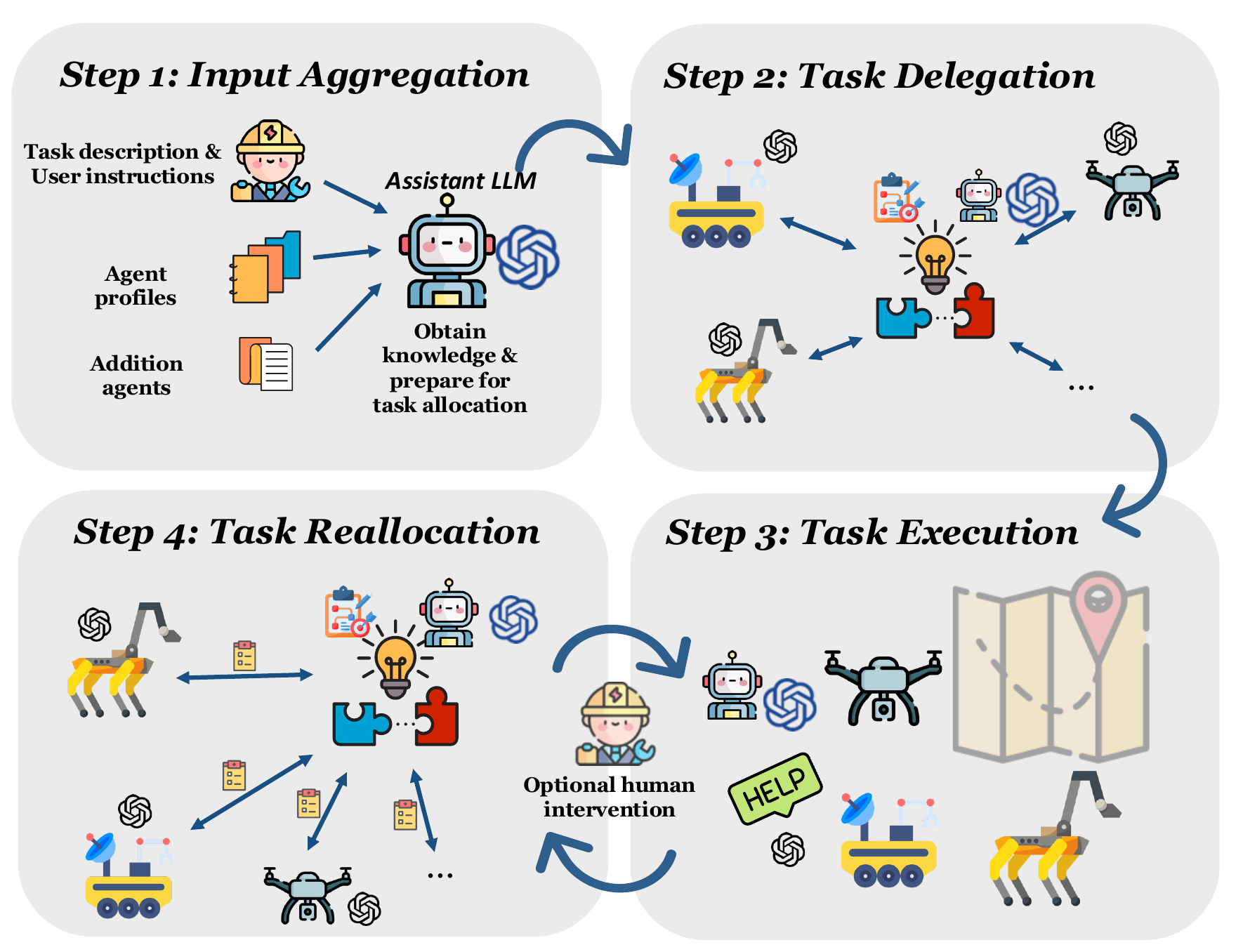}
  \caption{The workflow of HMCF. The system is first initialized by user input and robot configuration files, followed by task allocation and task execution. In case of failure of task execution of one or more agents, task reallocation is performed. User input can be involved in Step 3 and Step 4.}
  \label{fig:flow_chart}
\end{figure}

Figure \ref{fig:flow_chart} illustrates the overall workflow of HMCF, where the system operates sequentially through \textit{Step 1: Input Aggregation}, \textit{Step 2: Task Delegation and Verification}, \textit{Step 3: Task Execution}, and \textit{Step 4: Task Reallocation}. Upon initiating a new session (\textit{Step 1}), the system loads descriptions and profiles of all available robots, This includes specifications such as the robot type (e.g., wheeled, legged), current battery levels (e.g., 80\% charged), and traversability (e.g., capable of climbing stairs or navigating rough terrain). To enhance system flexibility and extend its general applicability, retrieval augmented generation (RAG) is incorporated, allowing new robot agents to be added by simply uploading their specification files, including names and descriptions (see Section 3.3 for more details). Furthermore, users may upload supplementary materials—such as robot manuals, datasheets, or user feedback—to build the knowledge base for newly added robots, improving the effectiveness of multi-robot collaboration during task execution.

After the \textit{human user} inputs task descriptions and instructions, the \textit{assistant LLM} agent constructs the knowledge base from the provided information. Following the completion of \textit{Step 1}, sub-tasks are assigned to each \textit{robot agent} based on their capabilities and the task requirements. Each robot then uses its own LLM agent to verify whether it can execute the assigned tasks and communicates the results back to the \textit{assistant LLM} agent. If task allocation is approved, the system proceeds to \textit{Step 3}, where agents begin executing their tasks. If not, the robots request task reallocation from the \textit{assistant LLM} agent.

During \textit{Step 3: Task Execution}, if a robot encounters an issue (e.g., unsuitable terrain or low battery), it signals an exception to the assistant LLM agent. The system then requests updates from all agents regarding their status, including location, remaining battery life, and task progress. Based on this information, the assistant LLM agent reallocates tasks as necessary (\textit{Step 4}). Each agent subsequently verifies the new tasks to ensure alignment with its capabilities before proceeding with execution.

In addition to automated task allocation, the system supports human intervention through a human-in-the-loop mechanism, enabling users to issue commands either as a broadcast to all robots or directly to specific agents. In broadcast mode, all robot agents and the assistant LLM agent receive and respond to commands, ensuring system-wide transparency. Once all LLM agents confirm the human instructions, each robot's agent verifies whether the reallocated tasks are feasible for the respective robot before execution. For direct control, users can command individual robots either through a group chat, where the instructions are visible to all agents and considered in future task allocations, or via a private chat, where only the targeted agent receives the command, maintaining confidentiality from other agents.

The chat history for each agent is logged to maintain context for future interactions. After each task allocation, the assistant LLM agent compiles a summary of the entire chat history and stores this summary for future sessions. When task reallocation is necessary, the assistant LLM agent refers to the summarized task history, which contains the assigned tasks and corresponding agents, instead of the complete chat transcript. This method reduces the reliance on lengthy text inputs and focuses on abstracted, essential information, thereby reducing LLM hallucination and enhancing the accuracy of task allocation.

\subsection{Human-Robots Interactive Interface} \label{section:ui}
\begin{figure}[t]
    \centering
    \includegraphics[width=\linewidth]{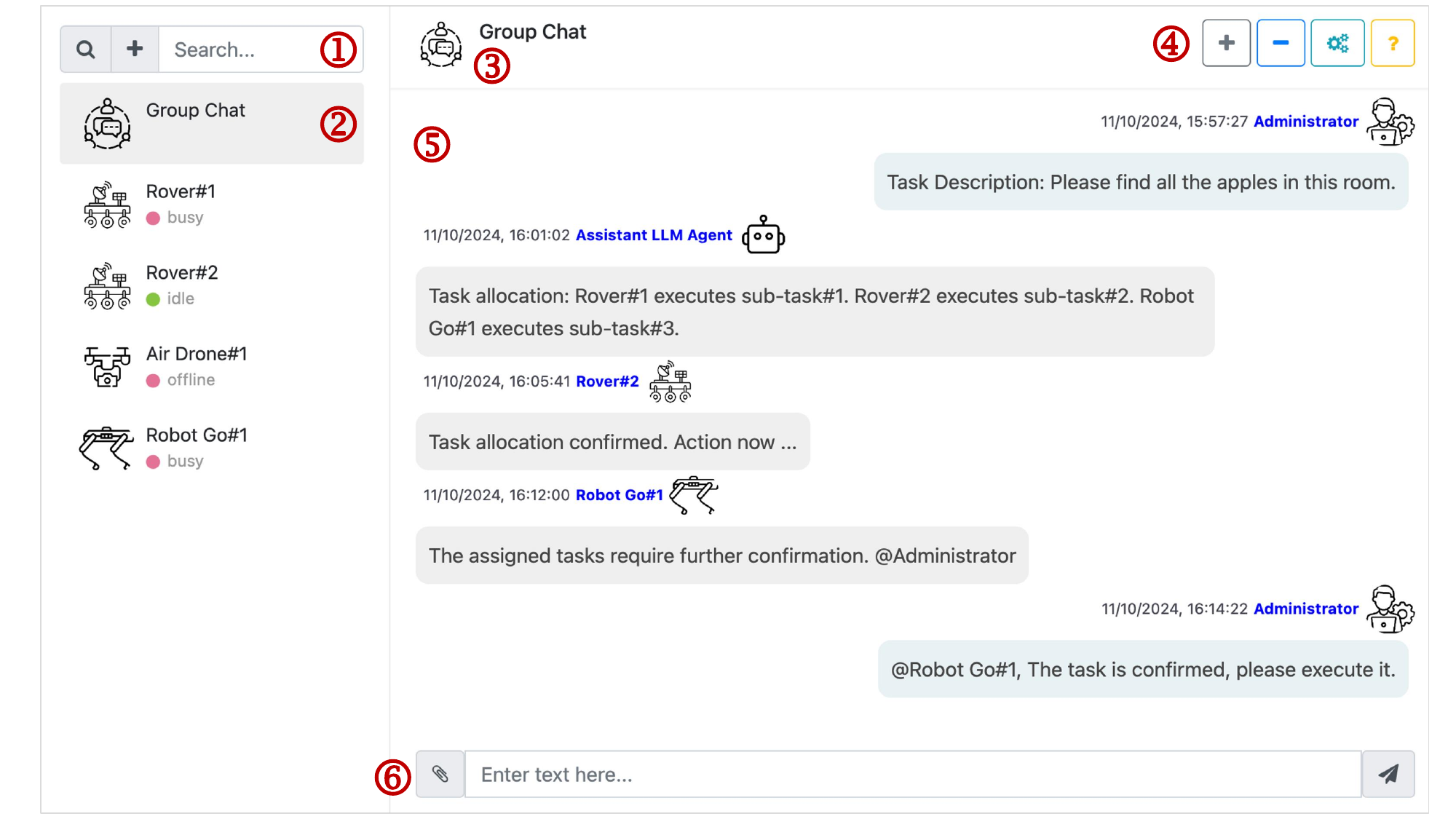}
    \caption{Graphical User Interface. The numbered regions represent 1) adding and querying robots, 2) robots chat list, 3) information of current chat, 4) configuration of cooperation group, 5) main chat window, and 6) user input box.} 
    \label{fig:gui}
\end{figure}

To empower human users with effective oversight capabilities, we developed a user-centric interactive interface aimed at facilitating human engagement in managing and supervising tasks across multiple heterogeneous robots. As shown in Figure \ref{fig:gui}, the design incorporates the human-in-the-loop (HITL) mechanism~\cite{wu2022survey}, allowing users to take an active role in both monitoring and decision-making processes. This interactive setup not only increases transparency in robot operations but also provides an intuitive method for system management, simplifying the supervision of complex multi-robot collaborations, even for non-experts. The interface is implemented as a web-based application, making it compatible with various platforms.

As illustrated in Figure \ref{fig:gui}, the user can initiate interactions by typing a task and environment description and sending it to the group chat, which consists of predefined robot agents. The interface will then display responses from both the assistant LLM agent and the robot agents within the chat window. Additionally, conversations about task reallocation and exceptions are shown in the same window, enabling users to monitor task execution in real-time.

In scenarios requiring control of a single robot, users can send commands by typing \texttt{@AgentName}, followed by the instructions, into the group chat. Only the specified agent will respond to the command, while the conversation remains visible to all agents. For manual task allocation or direct inquiries, users can select the desired robot from the list displayed on the left side of the window, initiating a one-on-one chat with the chosen robot.

\subsection{LLM-Augmented Agent}
In this project, we utilized OpenAI’s Assistant GPT API\footnote{https://platform.openai.com/docs/assistants/overview}, which differs from the standard API in that an assistant has pre-defined instructions and can leverage models, tools, and files to respond to user queries. Specifically, we employed the model’s multimodal capabilities and \textbf{\textit{file\_search}} functionality. The multimodal feature was used to convert visual information from robots into textual data, facilitating collaborative tasks and communication between robots. The file search function is a Retrieval-Augmented Generation (RAG)~\cite{lewis2020retrieval} method integrated into the ChatGPT-4 API. It works by retrieving relevant data or documents related to a given problem or task and using them as context for the language model. In our project, this capability was utilized to configure robots efficiently by uploading and reading their manuals. In the one-to-one channel (designed for individual communication with a robot), any file uploaded by the user is automatically incorporated into that specific agent's knowledge base. When responding to queries, the agent will retrieve and ground its answers using the background knowledge from the uploaded documents.

We used the GPT-4o-2024-08-06 model, which supports structured outputs with a context window of 128,000 tokens and a maximum output of 16,384 tokens\footnote{https://platform.openai.com/docs/models/gpt-4o}. Although this GPT-4o-2024-08-06 model supports a large context window, challenges arose with the occurrence of LLM hallucinations as the amount of communication and the number of robots increased. To mitigate this, similar to human input, we designed an "@" function, where the \textit{assistant LLM} agent explicitly assigns tasks to specific robots by tagging them. While the entire conversation is visible in the chatbox for human oversight, each robot’s LLM agent processes only the content tagged with its identity to avoid information overload, which could lead to hallucinations. The task allocation is presented in the following format in the chat: \texttt{"- [RobotName\_1] has been assigned [TaskName\_a]...}". The corresponding backend data format is structured as: \texttt{"@[RobotName\_1] Your task is [TaskName\_a]. EOF ..."}

As discussed in the previous section \ref{section:ui}, human users can also utilize the "\texttt{@}" function to assign specific tasks to one or multiple robots by explicitly tagging them, which could reduce hallucination issues caused by long context windows.

\subsection{Robot Deployments}
Our framework includes a base class, \texttt{Robot}, which provides interfaces for issuing robot-specific instructions. New robot types can be easily integrated into the system by inheriting from the \texttt{Robot} class and implementing hardware-specific functions. Our system can support a variety of robots, but for our real-world deployments, we selected only these three types of robots: a small rover, an augmented rover, and a legged robot. Once the system is operational, the \textit{assistant LLM} selects the appropriate functions for each robot to execute its assigned tasks. The \textit{assistant LLM} distributes high-level commands to the robot agent, such as searching area A. The robot agent then translates these commands into specific executable actions, such as moving forward, backward, or jumping.

For both wheeled and legged robots, we have defined a \newline\texttt{move\_to(coordinate)} function, which directs the robot to move to a specified position. In addition to this, platform-specific functions are implemented: the legged robot, for example, can jump using \texttt{jump\_upward()} and can ascend or descend stairs by calling \texttt{climb\_up()} or \texttt{climb\_down()} respectively. After these functions are executed, the low-level controllers translate them into the necessary actions for the robot. Should any unexpected events occur during task execution, the assistant LLM is alerted via exceptional messages. Additionally, all robots are equipped with \texttt{get\_status()} and \texttt{get\_task\_progress()} functions, which regularly send updates on the robot’s status and task progress to both the assistant LLM and the user. This information is used to inform decisions regarding task allocation.


\section{Experiments and Evaluation} \label{exp}

Although numerous multi-robot collaboration frameworks exist, the variability among different robots, platforms, and real-world environments poses challenges in establishing a standardized benchmark and uniform evaluation metrics for robot team collaboration. Therefore, we conduct the evaluation of our proposed framework in two separate parts. First, we assess task allocation capabilities in the simulated environment {BEHAVIOR-1K}~\cite{li2023behavior}, which provides a standardized platform for comparing different frameworks. In the second part, we configure our framework on a team of real heterogeneous robots, enabling them to perform various tasks across different environments. Testing in a simulation environment demonstrates the effectiveness of our framework for multi-robot task allocation, while real-world deployment shows its feasibility for practical applications.

\subsection{Simulation and Performance Analysis} \label{simulation}

\paragraph{\textbf{Testbed}}
To evaluate the task allocation capabilities of our framework, we used the {BEHAVIOR-1K}~\cite{li2023behavior} benchmark, a comprehensive simulation environment designed to replicate real-world tasks for robots. BEHAVIOR-1K consists of 1,000 everyday household activities grounded in human needs, covering a wide variety of scenes, objects, and conditions. It features a large-scale dataset with commonsense knowledge, predicate logic definitions for tasks, and over 9,000 object models with detailed physical and semantic properties.

We selected 5 distinct scenes from the BEHAVIOR-1K platform to evaluate our framework and baseline models. In these 5 scenes, \textbf{S1 to 5} correspond to the following environments: house, store, restaurant, office, and garden.

Each scene contains 10 different tasks. All 10 tasks require teams of three or more heterogeneous robots, including legged robots, wheeled robots, arm robots, and drones, to collaborate on tasks that require collaboration among heterogeneous robots. These tasks involve complex operations, such as jointly retrieving an object from one location and delivering it to another location, which may require actions like opening a door or accessing hard-to-reach areas. Each task in these scenes was executed 10 times to ensure consistent and reliable results.

To address the variability in task difficulty across the different scenes, we sought to minimize discrepancies in the minimum average number of steps required to complete each task, aiming for a consistent comparison across the five scenes. However, some variation still remains due to the inherent complexity of each task. In the selected five scenes, the difficulty levels of the scenes range from 1 to 5, with S1 being the easiest (difficulty level 1) and S5 the most challenging (difficulty level 5). These difficulty levels were determined based on the minimum number of steps required to complete the tasks in each scene, with scenes requiring fewer steps considered easier and those needing more steps classified as more difficult.

\paragraph{\textbf{Compared Models}}

To evaluate the performance of our model, we compared our approach with 3 traditional reinforcement learning based methods and 2 LLM-based multi-robot collaboration models:

\begin{itemize}
    \item \textbf{RL-VMC~\cite{haarnoja2018soft}}: A visuomotor control approach that leverages Soft Actor-Critic (SAC) to process image inputs and produce low-level joint control commands for robotic actions.
    
    \item \textbf{RL-Prim~\cite{schulman2017proximal}}: A reinforcement learning method built on Proximal Policy Optimization (PPO), which utilizes action primitives (such as pick, place, and navigate) derived from a sampling-based motion planner to execute tasks.
    
    \item \textbf{RL-Prim.Hist~\cite{li2023behavior}}: An extension of RL-Prim that integrates a history of the last three observations, allowing the agent to better distinguish between similar states and improve decision-making.
        
    \item \textbf{HMAS-2~\cite{chen2024scalable}}: A centralized system where a single LLM is responsible for planning actions for all robots, improving the efficiency of task collaboration in multi-agent setups.
    
    \item \textbf{DMRS-2D~\cite{zhang2023building}}: A decentralized communication framework where each robot is paired with its own LLM agent. Task planning and execution are performed through iterative rounds of communication, with each agent processing raw sensory observations to solve multi-objective tasks across embodied environments.

\end{itemize}

\begin{figure}[t]
  \centering
  \includegraphics[width=\linewidth]{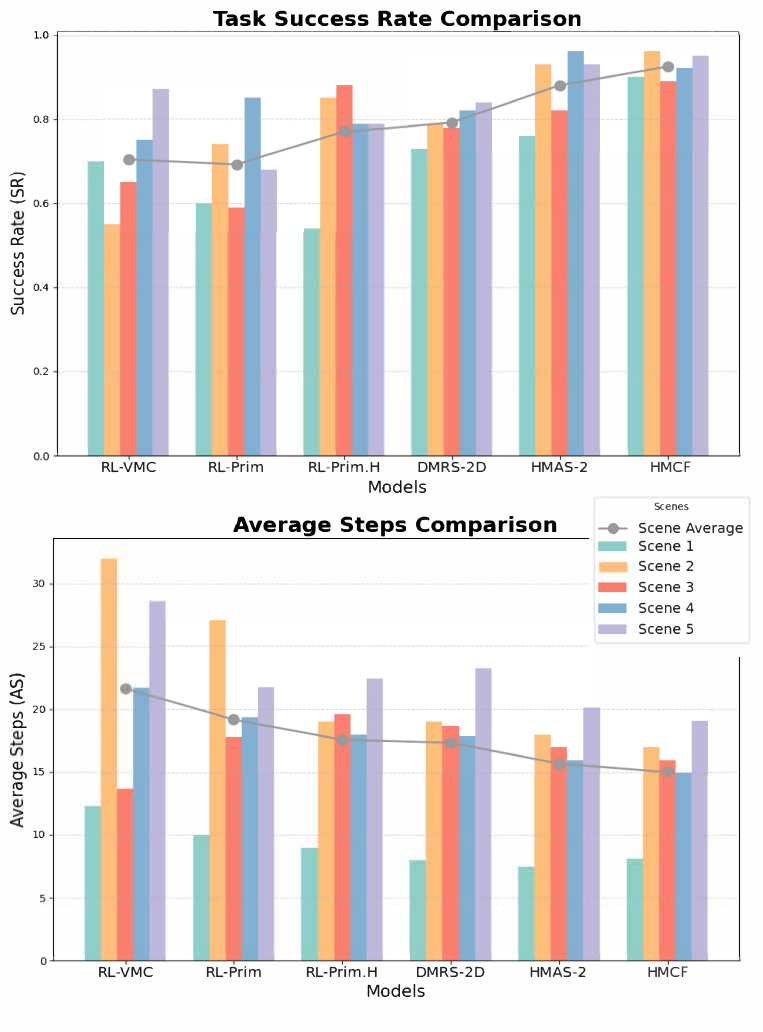}
  \caption{Comparison of Different Models Performance. Scenes 1 to 5 are house, store, restaurant, office, and garden, respectively.}
  \label{fig:comparison}
\end{figure}

To ensure fairness when comparing our framework to others without human involvement, in the simulation environment, the human supervisor's role is limited to describing the task at the outset. During task reallocation, if the robot agent and the task allocation agent disagree, they will request human intervention by tagging the supervisor in the group chat. The human supervisor can only respond once with 'yes' or 'no' and cannot provide any additional information that could aid the agents in obtaining more details. The one response from the human user is also counted in the execution steps. Since humans, as global observers in task execution, generally possess a correct understanding of these simple tasks, we do not consider the difference between humans and whether the supervisor has specialized knowledge or expertise in the simulated environment at this stage. When the agent requests assistance from a human, the supervisor will provide the correct answer to the agent only once. The human supervisor in these experiments is a robot expert familiar with the BEHAVIOR-1K platform, and the feedback provided was all correct. It is also important to note that, unlike our LLM-based framework, RL approaches do not utilize text data for task descriptions or interactions, making the comparison focused solely on task execution performance.


\paragraph{\textbf{Metrics}}
To evaluate the performance of each model, we utilized two primary metrics: Success Rate (SR) and Average Steps (AS). Success Rate (SR) is defined as the proportion of tasks successfully completed within the given constraints across the different scenes. It provides an indication of the model's ability to consistently accomplish tasks. Average Steps (AS) measures the average number of steps taken by each model to complete a task. This metric captures the models' efficiency in task execution, where a lower number of steps indicates higher efficiency. 

\paragraph{\textbf{Results}}

The results are summarized in Table \ref{tab:performance}, and the visualizations in Figure \ref{fig:comparison} provide a comparison of the models' success rates across the scenes.

Among the evaluated models, HMCF consistently demonstrated the best overall performance, achieving the highest average SR (0.924) and the lowest average AS (15.04). In particular, HMCF outperformed all other models across most scenes, while maintaining lower average steps in these scenes compared to others. This is clearly depicted in the comparison chart for HMCF, where it shows superior performance across all scenes compared to the average success rates.

The next best-performing model was HMAS-2, which achieved a strong average SR of 0.88 and the second-lowest average AS of 15.72. This model showed notable success in Scene 4 (office), where it achieved the highest SR (0.96) among all models. The comparison chart for HMAS-2 highlights its close performance to HMCF across multiple scenes. DMRS-2D and RL-Prim.H also performed reasonably well, with average SRs of 0.792 and 0.77, respectively. However, their performance in terms of average steps was slightly higher compared to HMCF and HMAS-2, indicating less efficiency in task execution. The comparison charts for these models show a more mixed performance, particularly in Scene 1 (house), where they fall behind.

The RL-Prim and RL-VMC models showed lower overall success rates, with RL-Prim achieving an average SR of 0.692 and RL-VMC scoring 0.704. Both models also had higher average steps (19.18 and 21.66, respectively), suggesting less effective performance in comparison to other models. The comparison charts for these models clearly illustrate their inconsistencies across the scenes, especially when compared to the average success rates.

\begin{table*}[htbp]
    \caption{Performance Comparison of Different Models on Task Success Rate (SR) and Average Steps (AS) Across Multiple Scene.}
  \centering
    \setlength{\tabcolsep}{3mm}
    \begin{tabular}{ccccccccccccc}
\hline
          & \multicolumn{2}{c}{Scene 1} & \multicolumn{2}{c}{Scene 2} & \multicolumn{2}{c}{Scene 3} & \multicolumn{2}{c}{Scene 4} & \multicolumn{2}{c}{Scene 5} & \multicolumn{2}{c}{\textbf{Average}}\\
    Models & SR  & AS & SR  & AS & AR  & AS & SR  & AS & SR  & AS & SR  & AS\\

\hline
   RL-VMC & 0.70  & 12.3  & 0.55  & 32  & 0.65  & \textbf{13.7} & 0.75  & 21.7 & 0.87  & 28.6 & 0.704  & 21.66 \\
   RL-Prim & 0.60  & 10  & 0.74  & 27  & 0.59  & 17.8 & 0.85  & 19.4 & 0.68  & 21.7 & 0.692  & 19.18 \\
   RL-Prim.H & 0.54  & 9   & 0.85  & 19  & 0.88  & 19.6 & 0.79  & 18  & 0.79  & 22.4 & 0.77  & 17.6 \\
   DMRS-2D  & 0.73  & \textbf{8}   & 0.79  & 19  & 0.78  & 18.7 & 0.82  & 17.9 & 0.84  & 23.2 & 0.792  & 17.36 \\
   HMAS-2 & 0.76  & 7.5   & 0.93  & 18   & 0.82  & 17  & \textbf{0.96}  & 16  & 0.93  & 20.1 & 0.88  & 15.72 \\
   \textbf{HMCF} & \textbf{0.90}  & 8.1   & \textbf{0.96}  & \textbf{17}   & \textbf{0.89}  & 16  & 0.92  & \textbf{15}  & \textbf{0.95}  & \textbf{19.1} & \textbf{0.92}  & \textbf{15.04} \\
\hline
    \end{tabular}%
     
  \label{tab:performance}%
\end{table*}%

\begin{table*}[htbp]
    \caption{Ablation Study Results: Comparison of Scene Success Rates (SR) and Average Steps (AS)}
  \centering
    \setlength{\tabcolsep}{3mm}
    \begin{tabular}{ccccccccccccc}
\hline
          & \multicolumn{2}{c}{Scene 1} & \multicolumn{2}{c}{Scene 2} & \multicolumn{2}{c}{Scene 3} & \multicolumn{2}{c}{Scene 4} & \multicolumn{2}{c}{Scene 5} & \multicolumn{2}{c}{ \textbf{Average}}\\
    Models & SR  & AS & SR  & AS & SR  & AS & SR  & AS & SR  & AS & SR  & AS\\
\hline
   \textbf{HMCF} & \textbf{0.90}  & \textbf{8.1}   & \textbf{0.96}  & \textbf{17}   & \textbf{0.89}  & \textbf{16}  & \textbf{0.92}  & \textbf{15}  & \textbf{0.95}  & \textbf{19.1} & \textbf{0.92}  & \textbf{15.04} \\
   HMCF-H     & 0.85  & 9.6   & 0.92  & 21   & 0.83  & 19    & 0.86  & 19    & 0.95  & 20.4   & 0.882  & 17.00  \\
   HMCF-H-V   & 0.65  & 12.5  & 0.80  & 24    & 0.78  & 35    & 0.77  & 27.5  & 0.83  & 22.1   & 0.766  & 24.22  \\
\hline
    \end{tabular}%
    \label{tab:ablation}
\end{table*}

\paragraph{\textbf{Ablation Study}}

\begin{figure}[t]
  \centering
  \includegraphics[width=\linewidth]{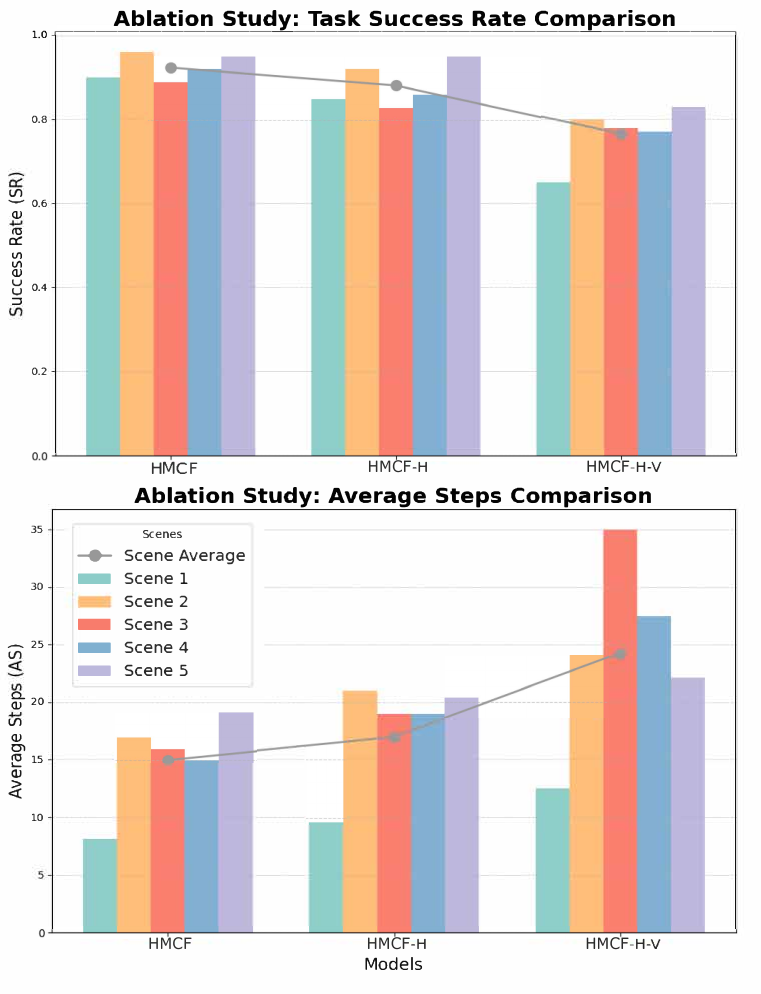}
  \caption{Ablation Study: Comparison of Three HMCF Models}
  \label{fig:ablation}
\end{figure}

In this ablation study, we compared the performance of our model, HMCF, with two variants: HMCF-H and HMCF-H-V. HMCF represents our full model, which incorporates a human-in-the-loop mechanism and specific heterogeneous agent task verification before task allocation. HMCF-H is a version of the model without the human-in-the-loop feature, while HMCF-H-V removes both the human-in-the-loop and the task verification by the heterogeneous agents. The performance of these models was evaluated across five scenes, focusing on Success Rate (SR) and Average Steps (AS), as shown in Table \ref{tab:ablation}. The results are further visualized through comparison charts, offering a comparative view of success rates across the scenes (Shown in Figure \ref{fig:ablation}).

HMCF performed the best overall, achieving the highest average SR (0.942) and the lowest average AS (15.04). This version achieved the highest average success rates (SR = 0.96) in Scene 2 (restaurant), and Scene 5 (garden) while also maintaining strong performance in the other scenes. The comparison chart highlights the consistent superiority of HMCF across all scenes, confirming that both the human-in-the-loop and the task verification mechanisms play crucial roles in achieving higher success rates and efficiency.

On the other hand, HMCF-H, the variant without the human-in-the-loop, showed a noticeable decline in performance. It had an average SR of 0.882 and an average AS of 17.00, indicating that the absence of human involvement in task oversight led to less efficient performance, particularly in Scene 2 (store) and Scene 3 (restaurant). This is evident in the comparison chart, where HMCF-H consistently underperforms when compared to HMCF.

HMCF-H-V, the version without both the human-in-the-loop and the task verification by the robot's agent, recorded the lowest performance among the three models. It achieved an average SR of 0.766 and the highest average AS of 24.22. The lack of both features resulted in a significant drop in task success rates, as well as higher inefficiency, as indicated by the increased steps required to complete tasks. The comparison chart for HMCF-H-V shows a substantial deviation from the success rates of the other models, particularly in Scene 3 (restaurant) and Scene 4 (office).

Overall, the ablation study highlights the critical importance of both the human-in-the-loop feature and task verification in improving task success rates and efficiency. The full HMCF model, which includes these features, clearly outperforms the other variants, confirming its superior effectiveness in multi-scene task completion.

\subsection{Real-World Deployment Cases}

\begin{figure}[t]
  \centering
  \includegraphics[width=\linewidth]{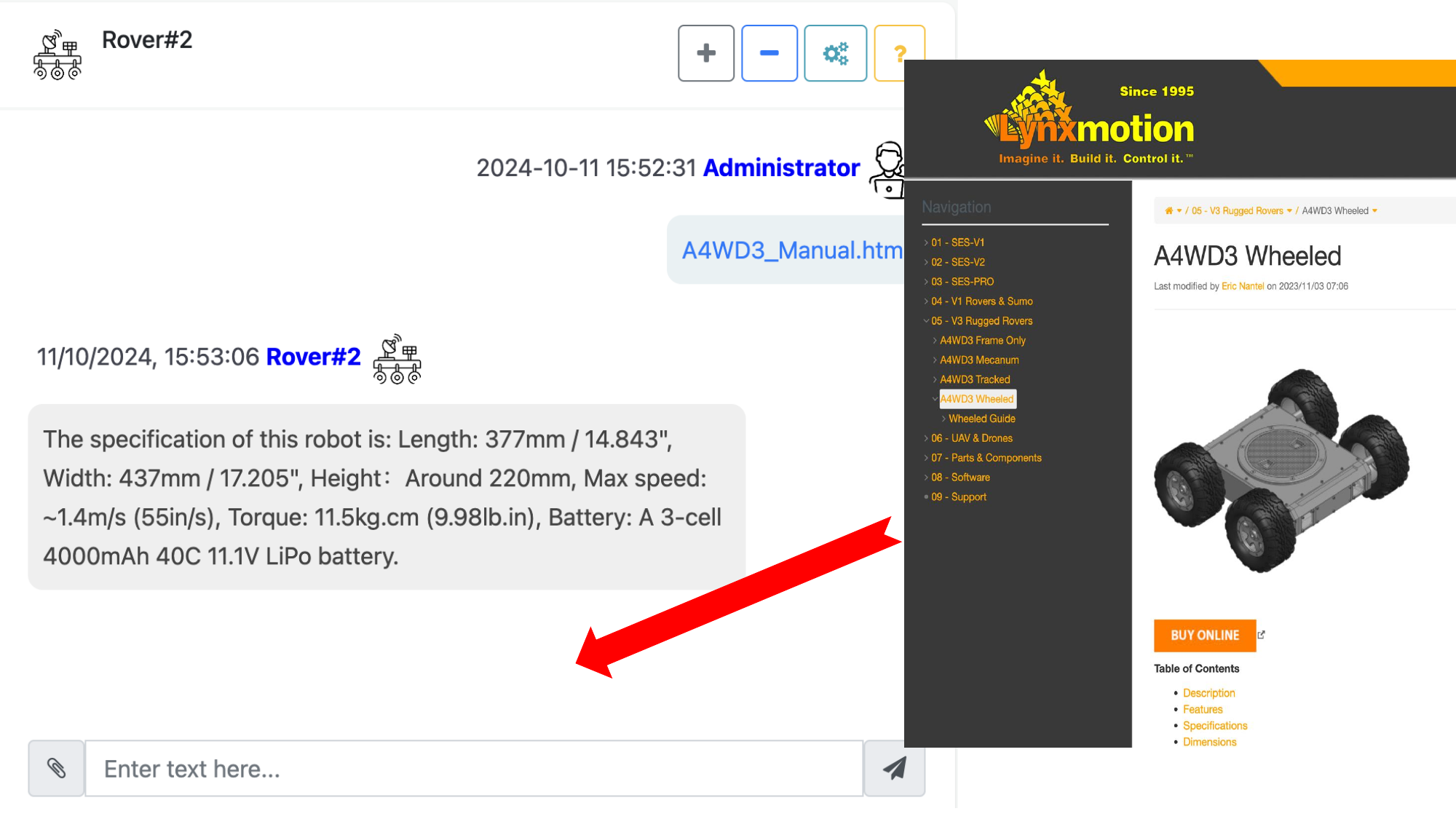}
  \caption{User-friendly configuration by providing a file of the user manual of the target robot.}
  \label{fig:RAG}
\end{figure}

\textbf{User-Friendly Configuration.} As we mentioned before, we can utilize the RAG capability of LLMs within the chatboxes, allowing users to configure robots easily in just a few steps. As shown in Figure \ref{fig:RAG}, by dragging the instruction manual of a Wheeled robot (Rover A4WD3) into the robot's chatbox, the LLM agent extracts key information from the robots, such as height, width, maximum speed, torque, and battery capacity. This information is essential for performing tasks in real-world environments. Users can also provide simple prompts to configure the robot's LLM agent further.

\textbf{Multimodal Capabilities based on LLMs.} In the traditional robotics field, when a robot encounters unforeseen obstacles, it can typically only issue a fault warning and capture images for human analysis~\cite{wang2024survey}. However, in our framework, with the multimodal capabilities of LLMs, the robot's LLM agent can analyze the situation directly using the photos taken by the camera and real-time status information sent back by the robot. 
For instance, when a rover robot encounters fallen branches in a dense forest and is unable to proceed, it can take a photo and send it to the LLM agent. The LLM agent could not only assess the situation and provide a solution back to the rover but also forward the analysis of the situation and key message to the \textit{assistant LLM}, which can coordinate other robots to assist in completing the task. Conversely, human users can also send images to the robot agents, instructing them to search for these items in the real environment.


\textbf{Task Allocation.}
To demonstrate the advantages of HMCF in the real world, we conducted experiments with a diverse team of heterogeneous robots. The group consisted of a small, cost-effective rover (Monsterborg), an augmented rover (A4WD3), and a legged robotic dog (Unitree GO2), each bringing different capabilities to the collaboration.

\begin{figure}[t]
  \centering
  \includegraphics[width=\linewidth]{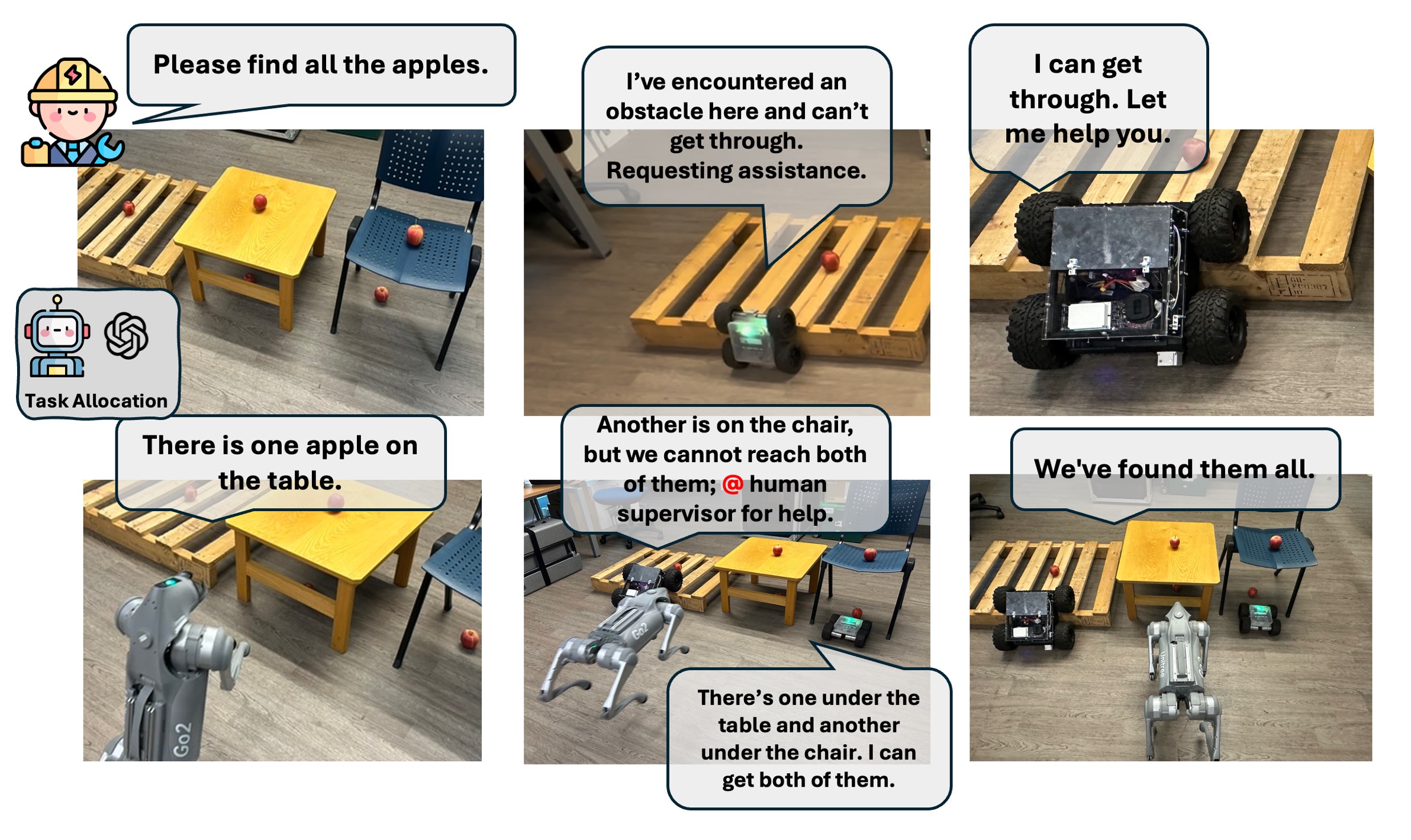}
  \caption{Real-World Deployment.}
  \label{fig:real}
\end{figure}

As illustrated in Figure \ref{fig:real}, a human supervisor initiated the task of locating all apples in the laboratory environment, prompting the robots to begin their search. The \textit{assistant LLMs} agent planned the tasks, prioritizing collision avoidance in the small space by first deploying the fastest and most agile rover to scan the area. When the smaller rover encountered an obstacle it could not overcome, the \textit{assistant LLM} agent reassigned the task to the larger rover, allowing the smaller one to explore other areas. Simultaneously, the robotic dog utilized its standing and jumping abilities to locate apples on the table and chairs. Since none of the robots could directly reach the apples, a request for assistance was sent to the human supervisor. Ultimately, all the apples were successfully found. This experiment demonstrates that HMCF could be effectively deployed in real-world scenarios, dynamically reallocating tasks by leveraging the diverse capabilities of heterogeneous robots.

\section{Discussion} \label{dis}

Our experimental results demonstrate that the HMCF framework consistently outperforms traditional reinforcement learning and other LLM-based frameworks across multiple simulation scenarios. In simulated environments, HMCF achieved a 92.4\% success rate, requiring significantly fewer steps for task completion than baseline models. The human-in-the-loop mechanism improved task success rates by 4.76\%, while decentralized task verification reduced execution errors. These findings validate the effectiveness of human oversight and heterogeneous robot collaboration in enabling efficient and reliable multi-robot coordination.

Despite its strengths, HMCF has limitations that warrant further research. Scalability remains a challenge in large-scale deployments involving dozens or hundreds of robots. While the current architecture performs well in smaller scenarios, optimizing communication overhead and minimizing latency will be critical for broader applications. Another limitation is the reliance on cloud-based LLMs, where communication between local robot systems and cloud-hosted LLM agents is constrained by signal transmission delays. Ensuring reliable connectivity in complex, real-world environments remains a key hurdle for large-scale deployment.

Additionally, the framework assumes continuous and stable communication between robots, which may not always be feasible in environments with poor connectivity. Future improvements should focus on robust communication protocols and fallback strategies to maintain seamless operations under adverse conditions.

To enhance human-robot interaction, we plan to conduct large-scale user studies involving non-expert users. These studies will help identify usability challenges, enabling refinements to the system interface and interaction mechanisms. By incorporating feedback from a diverse user base, we aim to make the system more intuitive and accessible, reducing reliance on specialized knowledge and improving overall user experience.

\section{Conclusion} \label{con}
This paper introduced \textbf{HMCF}, a human-in-the-loop multi-robot collaboration framework that leverages LLMs to address key challenges in multi-robot systems. Our framework enhances scalability, adaptability, and reliability by offering a human-robot interaction interface and integrating both human oversight and task verification. Experimental results showed that HMCF outperformed traditional reinforcement learning and LLM-based task allocation methods, achieving higher task success rates, efficiency, and scalability across both simulated and real-world environments. Future work will explore its deployment across diverse industrial domains.
\bibliographystyle{IEEEtran}
\bibliography{ref}

\begin{thebibliography}{10}
\providecommand{\url}[1]{#1}
\csname url@samestyle\endcsname
\providecommand{\newblock}{\relax}
\providecommand{\bibinfo}[2]{#2}
\providecommand{\BIBentrySTDinterwordspacing}{\spaceskip=0pt\relax}
\providecommand{\BIBentryALTinterwordstretchfactor}{4}
\providecommand{\BIBentryALTinterwordspacing}{\spaceskip=\fontdimen2\font plus
\BIBentryALTinterwordstretchfactor\fontdimen3\font minus \fontdimen4\font\relax}
\providecommand{\BIBforeignlanguage}[2]{{%
\expandafter\ifx\csname l@#1\endcsname\relax
\typeout{** WARNING: IEEEtran.bst: No hyphenation pattern has been}%
\typeout{** loaded for the language `#1'. Using the pattern for}%
\typeout{** the default language instead.}%
\else
\language=\csname l@#1\endcsname
\fi
#2}}
\providecommand{\BIBdecl}{\relax}
\BIBdecl

\bibitem{mitchell2022review}
D.~Mitchell, J.~Blanche, S.~Harper, T.~Lim, R.~Gupta, O.~Zaki, W.~Tang, V.~Robu, S.~Watson, and D.~Flynn, ``A review: Challenges and opportunities for artificial intelligence and robotics in the offshore wind sector,'' \emph{Energy and AI}, vol.~8, p. 100146, 2022.

\bibitem{ghassemi2022multi}
P.~Ghassemi and S.~Chowdhury, ``Multi-robot task allocation in disaster response: Addressing dynamic tasks with deadlines and robots with range and payload constraints,'' \emph{Robotics and Autonomous Systems}, vol. 147, p. 103905, 2022.

\bibitem{wu2024proof}
E.~W. Wu, M.~Jurt, B.~Holden, and Y.~Jin, ``Proof of location verification towards trustworthy collaborative multi-vendor robotic systems,'' in \emph{2024 IEEE International Conference on Industrial Technology (ICIT)}.\hskip 1em plus 0.5em minus 0.4em\relax IEEE, 2024, pp. 1--8.

\bibitem{chen2023agentverse}
W.~Chen, Y.~Su, J.~Zuo, C.~Yang, C.~Yuan, C.~Qian, C.-M. Chan, Y.~Qin, Y.~Lu, R.~Xie \emph{et~al.}, ``Agentverse: Facilitating multi-agent collaboration and exploring emergent behaviors in agents,'' \emph{arXiv preprint arXiv:2308.10848}, vol.~2, no.~4, p.~6, 2023.

\bibitem{chen2024scalable}
Y.~Chen, J.~Arkin, Y.~Zhang, N.~Roy, and C.~Fan, ``Scalable multi-robot collaboration with large language models: Centralized or decentralized systems?'' in \emph{2024 IEEE International Conference on Robotics and Automation (ICRA)}.\hskip 1em plus 0.5em minus 0.4em\relax IEEE, 2024, pp. 4311--4317.

\bibitem{mandi2024roco}
Z.~Mandi, S.~Jain, and S.~Song, ``Roco: Dialectic multi-robot collaboration with large language models,'' in \emph{2024 IEEE International Conference on Robotics and Automation (ICRA)}.\hskip 1em plus 0.5em minus 0.4em\relax IEEE, 2024, pp. 286--299.

\bibitem{ye2023cognitive}
H.~Ye, T.~Liu, A.~Zhang, W.~Hua, and W.~Jia, ``Cognitive mirage: A review of hallucinations in large language models,'' \emph{arXiv preprint arXiv:2309.06794}, 2023.

\bibitem{kim2024survey}
Y.~Kim, D.~Kim, J.~Choi, J.~Park, N.~Oh, and D.~Park, ``A survey on integration of large language models with intelligent robots,'' \emph{Intelligent Service Robotics}, vol.~17, no.~5, pp. 1091--1107, 2024.

\bibitem{li2021survey}
Z.~Li, L.~Shi, A.~I. Cristea, and Y.~Zhou, ``A survey of collaborative reinforcement learning: interactive methods and design patterns,'' in \emph{Proceedings of the 2021 ACM Designing Interactive Systems Conference}, 2021, pp. 1579--1590.

\bibitem{li2024towards}
Z.~Li, V.~Yazdanpanah, S.~Sarkadi, Y.~He, E.~Shafipour, and S.~Stein, ``Towards citizen-centric multiagent systems based on large language models,'' in \emph{Proceedings of the 2024 International Conference on Information Technology for Social Good}, 2024, pp. 26--31.

\bibitem{wang2024survey}
L.~Wang, C.~Ma, X.~Feng, Z.~Zhang, H.~Yang, J.~Zhang, Z.~Chen, J.~Tang, X.~Chen, Y.~Lin \emph{et~al.}, ``A survey on large language model based autonomous agents,'' \emph{Frontiers of Computer Science}, vol.~18, no.~6, pp. 1--26, 2024.

\bibitem{guo2024large}
T.~Guo, X.~Chen, Y.~Wang, R.~Chang, S.~Pei, N.~V. Chawla, O.~Wiest, and X.~Zhang, ``Large language model based multi-agents: A survey of progress and challenges,'' \emph{arXiv preprint arXiv:2402.01680}, 2024.

\bibitem{liang2020emergent}
P.~P. Liang, J.~Chen, R.~Salakhutdinov, L.-P. Morency, and S.~Kottur, ``On emergent communication in competitive multi-agent teams,'' \emph{arXiv preprint arXiv:2003.01848}, 2020.

\bibitem{li2023behavior}
C.~Li, R.~Zhang, J.~Wong, C.~Gokmen, S.~Srivastava, R.~Mart{\'\i}n-Mart{\'\i}n, C.~Wang, G.~Levine, M.~Lingelbach, J.~Sun \emph{et~al.}, ``Behavior-1k: A benchmark for embodied {AI} with 1,000 everyday activities and realistic simulation,'' in \emph{Conference on Robot Learning}.\hskip 1em plus 0.5em minus 0.4em\relax PMLR, 2023, pp. 80--93.

\bibitem{mozannar2024effective}
H.~Mozannar, J.~Lee, D.~Wei, P.~Sattigeri, S.~Das, and D.~Sontag, ``Effective human-ai teams via learned natural language rules and onboarding,'' \emph{Advances in Neural Information Processing Systems}, vol.~36, 2024.

\bibitem{li2023broader}
Z.~Li, M.~Jacobsen, L.~Shi, Y.~Zhou, and J.~Wang, ``Broader and deeper: A multi-features with latent relations bert knowledge tracing model,'' in \emph{European Conference on Technology Enhanced Learning}.\hskip 1em plus 0.5em minus 0.4em\relax Springer, 2023, pp. 183--197.

\bibitem{brohan2023rt}
A.~Brohan, N.~Brown, J.~Carbajal, Y.~Chebotar, X.~Chen, K.~Choromanski, T.~Ding, D.~Driess, A.~Dubey, C.~Finn \emph{et~al.}, ``Rt-2: Vision-language-action models transfer web knowledge to robotic control,'' \emph{arXiv preprint arXiv:2307.15818}, 2023.

\bibitem{li2023deep}
Z.~Li, ``Deep reinforcement learning approaches for technology enhanced learning,'' Ph.D. dissertation, Durham University, 2023.

\bibitem{shen2024language}
L.~Shen, W.~Tan, S.~Chen, Y.~Chen, J.~Zhang, H.~Xu, B.~Zheng, P.~Koehn, and D.~Khashabi, ``The language barrier: Dissecting safety challenges of llms in multilingual contexts,'' \emph{arXiv preprint arXiv:2401.13136}, 2024.

\bibitem{li2023sim}
Z.~Li, L.~Shi, J.~Wang, A.~I. Cristea, and Y.~Zhou, ``Sim-gail: A generative adversarial imitation learning approach of student modelling for intelligent tutoring systems,'' \emph{Neural Computing and Applications}, vol.~35, no.~34, pp. 24\,369--24\,388, 2023.

\bibitem{li2023towards}
Z.~Li, L.~Shi, Y.~Zhou, and J.~Wang, ``Towards student behaviour simulation: a decision transformer based approach,'' in \emph{International Conference on Intelligent Tutoring Systems}.\hskip 1em plus 0.5em minus 0.4em\relax Springer, 2023, pp. 553--562.

\bibitem{valmeekam2024planbench}
K.~Valmeekam, M.~Marquez, A.~Olmo, S.~Sreedharan, and S.~Kambhampati, ``Planbench: An extensible benchmark for evaluating large language models on planning and reasoning about change,'' \emph{Advances in Neural Information Processing Systems}, vol.~36, 2024.

\bibitem{li2024design}
Z.~Li, ``A design trajectory map of human-ai collaborative reinforcement learning systems: Survey and taxonomy,'' \emph{arXiv preprint arXiv:2405.10214}, 2024.

\bibitem{kannan2023smart}
S.~S. Kannan, V.~L. Venkatesh, and B.-C. Min, ``Smart-llm: Smart multi-agent robot task planning using large language models,'' \emph{arXiv preprint arXiv:2309.10062}, 2023.

\bibitem{li2024integrating}
Z.~Li, J.~Yang, J.~Wang, L.~Shi, and S.~Stein, ``Integrating lstm and bert for long-sequence data analysis in intelligent tutoring systems,'' \emph{arXiv preprint arXiv:2405.05136}, 2024.

\bibitem{li2024lbkt}
Z.~Li, J.~Yang, J.~Wang, L.~Shi, J.~Feng, and S.~Stein, ``Lbkt: a lstm bert-based knowledge tracing model for long-sequence data,'' in \emph{International Conference on Intelligent Tutoring Systems}.\hskip 1em plus 0.5em minus 0.4em\relax Springer, 2024, pp. 174--184.

\bibitem{zhang2023building}
H.~Zhang, W.~Du, J.~Shan, Q.~Zhou, Y.~Du, J.~B. Tenenbaum, T.~Shu, and C.~Gan, ``Building cooperative embodied agents modularly with large language models,'' \emph{arXiv preprint:2307.02485}, 2023.

\bibitem{martino2023knowledge}
A.~Martino, M.~Iannelli, and C.~Truong, ``Knowledge injection to counter large language model (llm) hallucination,'' in \emph{European Semantic Web Conference}.\hskip 1em plus 0.5em minus 0.4em\relax Springer, 2023, pp. 182--185.

\bibitem{hussein2018mixed}
A.~Hussein and H.~Abbass, ``Mixed initiative systems for human-swarm interaction: Opportunities and challenges,'' in \emph{2018 2nd Annual Systems Modelling Conference (SMC)}.\hskip 1em plus 0.5em minus 0.4em\relax IEEE, 2018, pp. 1--8.

\bibitem{wu2022survey}
X.~Wu, L.~Xiao, Y.~Sun, J.~Zhang, T.~Ma, and L.~He, ``A survey of human-in-the-loop for machine learning,'' \emph{Future Generation Computer Systems}, vol. 135, pp. 364--381, 2022.

\bibitem{lewis2020retrieval}
P.~Lewis, E.~Perez, A.~Piktus, F.~Petroni, V.~Karpukhin, N.~Goyal, H.~K{\"u}ttler, M.~Lewis, W.-t. Yih, T.~Rockt{\"a}schel \emph{et~al.}, ``Retrieval-augmented generation for knowledge-intensive nlp tasks,'' \emph{Advances in Neural Information Processing Systems}, vol.~33, pp. 9459--9474, 2020.

\bibitem{haarnoja2018soft}
T.~Haarnoja, A.~Zhou, P.~Abbeel, and S.~Levine, ``Soft actor-critic: Off-policy maximum entropy deep reinforcement learning with a stochastic actor,'' in \emph{International conference on machine learning}.\hskip 1em plus 0.5em minus 0.4em\relax PMLR, 2018, pp. 1861--70.

\bibitem{schulman2017proximal}
J.~Schulman, F.~Wolski, P.~Dhariwal, A.~Radford, and O.~Klimov, ``Proximal policy optimization algorithms,'' \emph{arXiv preprint arXiv:1707.06347}, 2017.

\end{thebibliography}

\end{document}